\documentclass[tinyml]{acmart}

\AtBeginDocument{%
  \providecommand\BibTeX{{%
    \normalfont B\kern-0.5em{\scshape i\kern-0.25em b}\kern-0.8em\TeX}}}

\setcopyright{rightsretained}
\copyrightyear{2022}
\acmYear{2022}

\usepackage{tikz}
\usepackage{subfigure}
\usepackage{multirow}
\usepackage{xspace}

\newcommand*\circled[1]{\tikz[baseline=(char.base)]{
            \node[shape=circle,draw,inner sep=0.1pt] (char) {#1};}}
\newcommand{\ouralg}{LDC\xspace}

\begin{document}

\title{A Brain-Inspired Low-Dimensional Computing Classifier for Inference on Tiny Devices}

\author{Shijin Duan}
\email{duan.s@northeastern.edu}
\affiliation{
 \institution{Northeastern University}
 \city{Boston}
 \state{MA}
 \country{USA}
}

\author{Xiaolin Xu}
\email{x.xu@northeastern.edu}
\affiliation{
 \institution{Northeastern University}
 \city{Boston}
 \state{MA}
 \country{USA}
}

\author{Shaolei Ren}
\email{sren@ece.ucr.edu}
\affiliation{
 \institution{UC Riverside}
 \city{Riverside}
 \state{CA}
 \country{USA}
}

\begin{abstract}
By mimicking brain-like cognition and exploiting parallelism, hyperdimensional computing (HDC) classifiers have been emerging as a lightweight framework to achieve efficient on-device inference. Nonetheless, they have two fundamental drawbacks --- heuristic training process and ultra-high dimension --- which result in sub-optimal inference accuracy and large model sizes beyond the capability of tiny devices with stringent resource constraints. In this paper, we address these fundamental drawbacks and propose a low-dimensional computing (\ouralg) alternative. Specifically, by mapping our \ouralg classifier into an equivalent neural network, we optimize  our model using a principled training approach. Most importantly, we can improve the inference accuracy while successfully reducing the ultra-high dimension of existing HDC models by orders of magnitude (e.g., 8000 vs. 4/64). We run experiments to evaluate our \ouralg classifier by considering different datasets for inference on tiny devices, and also implement different models on an FPGA platform for acceleration. The results highlight that our \ouralg classifier offers an overwhelming advantage over the existing brain-inspired HDC models and is particularly suitable for inference on tiny devices.
\end{abstract}

\maketitle

\section{Introduction}
\label{sec:introduction}

Deploying deep neural networks (DNNs) for on-device inference, especially on tiny Internet of Things (IoT) devices with stringent resource constraints, presents a key challenge \cite{DNN_NAS_HardwareNAS201_Benchmark_Rice_ICLR_2021_li2021hwnasbench,DNN_Compression_SongHan_ICLR_2016,DNN_Compression_Structured_YiranChen_NIPS_2016_10.5555/3157096.3157329}. This is due in part to the fundamental limitation of DNNs that involve intensive mathematical operators and computing beyond the capability of many tiny devices \cite{TinyML_MCUNet_TinyIoT_SongHan_NIPS_2020_NEURIPS2020_86c51678}.

More recently, inspired from the human-brain memorizing mechanism, hyperdimensional computing (HDC) for classification has been emerging  as a lightweight machine learning framework targeting inference on resource-constrained edge devices \cite{kanerva2009hyperdimensional,HDC_BinaryLearning_UCSD_DATE_2019_8714821}. In a nutshell, HDC classifiers mimic the brain cognition process by representing an object as a vector (a.k.a. \emph{hypervector}) with a very high dimension in the order of thousands or even more. They perform inference by comparing the similarities between the hypervector of a testing sample and a set of pre-built hypervectors representing different classes. Thus, with HDC, the conventional DNN inference process is essentially projected to parallelizable bit-wise operation in a hyperdimesional space. This offers several key advantages to HDC over its DNN counterpart, including high energy efficiency and low latency, and hence makes HDC classifiers potentially promising for on-device inference \cite{HDC_Survey_Review_IEEE_Circuit_Magazine_2020_9107175,imani2019quanthd,imani2021revisiting}. As a consequence, the set of studies on optimizing HDC classifier performance in terms of inference accuracy, latency and/or energy consumption has quickly expanding \cite{HDC_BinaryLearning_UCSD_DATE_2019_8714821,HDC_Survey_Review_IEEE_Circuit_Magazine_2020_9107175,10.1145/2591971.2591990,HDC_Combine_DensityEncoding_Random_Connected_NN_arXiv_kleyko2019density}.

Nonetheless, state-of-the-art (SOTA) HDC classifiers suffer from fundamental drawbacks that prevent their successful deployment for inference on \emph{tiny} devices. First and foremost, the \emph{hyperdimensional} nature of HDC means that each value or feature is represented by a hypervector with several thousands of or even more bits, which can easily result in a prohibitively large model size beyond the limit of typical tiny devices. Even putting aside the large HDC model size, parallel processing of a huge number of bit-wise operations associated with hypervectors is barely feasible on tiny devices, thus significantly increasing the inference latency. Furthermore, the energy consumption by processing hypervectors can also be a deal breaker for inference on tiny devices. In addition, another crucial drawback of HDC classifiers is the low inference accuracy resulting from the lack of a principled training approach. Concretely, the training process of an HDC classifier is extremely simple --- simply averaging over the hypervectors of labeled training samples to derive the corresponding class hypervectors. Although some heuristic techniques (e.g., re-training and regeneration \cite{HDC_NeuralAdaptation_UCI_SC_2021_10.1145/3458817.3480958,imani2019searchd,HDC_Survey_Review_IEEE_Circuit_Magazine_2020_9107175}) have been recently added, the existing HDC training process still lacks rigorousness and heavily relies on a trial-and-error process without systematic guidance as in the realm of DNNs.
In fact, even a well-defined loss function is lacking in the training of HDC
classifiers.

In this paper, we address the fundamental challenges in the existing brain-inspired HDC classifiers and propose a new ultra-efficient classification model based on \emph{low}-dimensional computing (\ouralg) for inference on resource-constrained tiny devices. Here, ``low'' is relative to the existing HDC model with a dimension of thousands or more.

First, we map the inference process of our \ouralg classifier into an equivalent neural network that includes a non-binary neural network followed by a binary neural network (BNN). Next, through the lens of this mapping, we optimize the neural network weights, from which we can extract low-dimensional vectors to represent object values and features for efficient inference. Most crucially, our \ouralg classifier eliminates the large hypervectors used in the existing HDC models, and utilizes optimized low-dimensional vectors with a much smaller size (e.g., 8000 vs. 4/64) to achieve even higher inference accuracy. Thus, compared to the existing SOTA HDC models, our \ouralg classifier can improve the inference accuracy and meanwhile dramatically reduce the model size, inference latency, and energy consumption by orders of magnitude.

We implement our \ouralg classifier on an FPGA platform under stringent resource constraints. The results show that, in addition to the improved accuracy (92.72\% vs. 87.38\%) on MNIST, our \ouralg classifier has a model size of \textbf{6.48 KB}, inference latency of \textbf{3.99 microseconds}, and inference energy of \textbf{64 nanojoules}, which are 100+ times smaller than a SOTA HDC model; for cardiotocography application, we achieve an accuracy of 90.50\%, with a model size of \textbf{0.32 KB}, inference latency of \textbf{0.14 microseconds}, and inference energy of \textbf{0.945 nanojoules}. The overwhelming advantage over the existing HDC models makes our \ouralg classifier particularly appealing for inference on tiny devices.

\section{Brain-Inspired HDC Classifiers}
\label{sec:background}

\begin{figure}[!t]
  \centering
  \includegraphics[width=\linewidth]{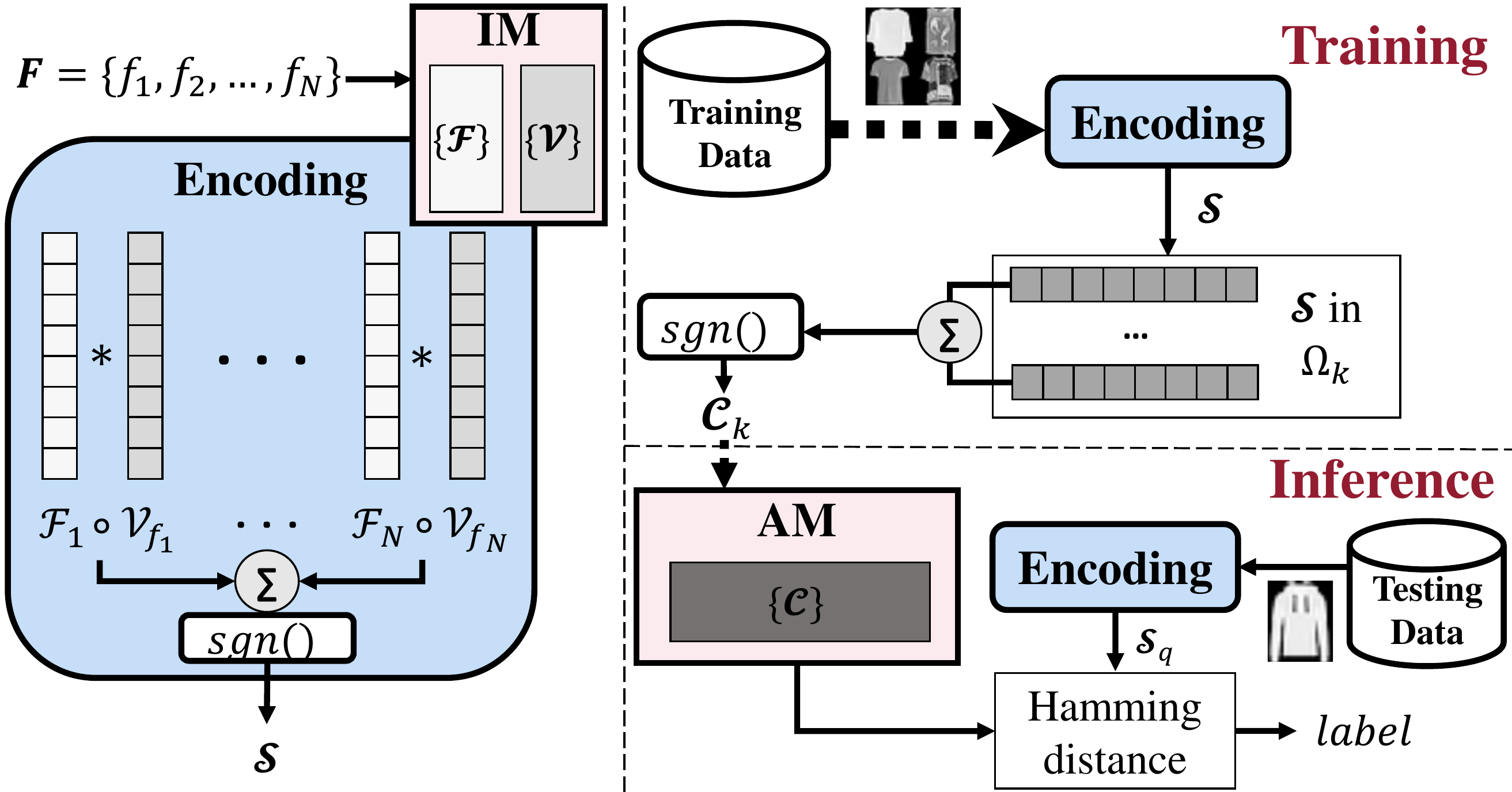}
  \caption{A brain-inspired HDC classifier. IM: item memory. AM: associative memory.}
  \label{fig:binary_HDC}
\end{figure}

We provide an overview of HDC models as illustrated in Figure~\ref{fig:binary_HDC}. Consider an input sample represented as a vector with $N$ features $\mathbf{F} = \{f_1,f_2,...,f_N\}$, where the value range for each feature is normalized and uniformly discretized into $M$ values, i.e., $f_i\in \{1,\cdots, M\}$ for $i=1,\cdots,N$. HDC encodes all the values, features, and samples as hyperdimensional bipolar vectors, e.g., $\mathcal{H}\in \{1,-1\}^{D=10,000}$, which also equivalently correspond to binary 0/1 bits for hardware efficiency \cite{HDC_BinaryLearning_UCSD_DATE_2019_8714821,HDC_Survey_Review_IEEE_Circuit_Magazine_2020_9107175}. In this paper, we also interchangeably use binary and bipolar if applicable. Note that the input vector $\mathbf{F} = \{f_1,f_2,...,f_N\}$
can represent raw features or extracted features (using,
e.g., neural networks and random feature map \cite{HDC_NeuralAdaptation_UCI_SC_2021_10.1145/3458817.3480958}).

There are four types of hypervectors in HDC models: \emph{value} hypervector $\mathcal{V}$ (representing the value of a feature), \emph{feature} hypervector $\mathcal{F}$ (representing the index/position of a feature), \emph{sample} hypervector $\mathcal{S}$ (representing a training/testing sample), and \emph{class} hypervector $\mathcal{C}$ (representing  a class). 

To measure the similarity between two hypervectors, there are two commonly used distances --- normalized Hamming and cosine, which are mutually equivalent as shown in the appendix. In this paper, we use the normalized Hamming distance defined as $Hamm(\mathcal{H}_1, \mathcal{H}_2) = \frac{|\mathcal{H}_1\neq \mathcal{H}_2|}{D}$. If two  hypervectors $\mathcal{H}_1$ and $\mathcal{H}_2$ have a normalized Hamming distance of 0.5, they are considered orthogonal. In a hyperdimensional space, two randomly-generated hypervectors are almost orthogonal \cite{HDC_inMemory_HDC_NatureElectronics_2020_IBM_ETHZ,HDC_Survey_Review_IEEE_Circuit_Magazine_2020_9107175}.

{\textbf{Generating value and feature hypervectors:}} 
In a typical HDC model \cite{kanerva2009hyperdimensional,HDC_Survey_Review_IEEE_Circuit_Magazine_2020_9107175,HDC_FastFPGA_UCSD_FPGA_2019_10.1145/3289602.3293913}, the value and feature hypervectors are randomly generated in advance and remain unchanged throughout the training and inference process. Most commonly, $N$ feature hypervectors $\mathcal{F}$ are randomly generated to keep mutual orthogonality (e.g., randomly sampling in the hyperdimensional space or rotating one random hypervector), whereas $M$ value hypervectors $\mathcal{V}$ are generated to preserve their value correlations (e.g., flipping a certain number of bits from one randomly-generated value hypervector) \cite{HDC_BinaryLearning_UCSD_DATE_2019_8714821,HDC_Survey_Review_IEEE_Circuit_Magazine_2020_9107175,kanerva2009hyperdimensional}. As a result, the Hamming distance between any two feature hypervectors is approximately 0.5, while the Hamming distance between two value hypervectors denoting normalized feature values of $i,j\in \{1,\cdots, M\}$ is $Hamm(\mathcal{V}_{i},\mathcal{V}_{j}) \approx \frac{|i-j|}{2(M-1)}$. 

{\textbf{Encoding:}} An input sample is encoded as a sample hypervector by fetching the pre-generated value and feature hypervectors from the item memory (IM).\footnote{The study \cite{HDC_NeuralAdaptation_UCI_SC_2021_10.1145/3458817.3480958} applies
 random feature map for feature extraction \cite{ML_RandomFeatureMap_Kernel_Berkeley_NIPS_2007_10.5555/2981562.2981710} and then directly binarizes the extracted features as the encoded hypervector. Nonetheless, we can also use value hypervectors
 and feature hypervectors to encode the \emph{non}-binary features
 extracted via random feature map, and this can preserve more information in
 the extracted features than direct binarization.}  Specifically, by combining each feature hypervector with its corresponding value hypervector, the encoding output for an input sample is given by 
\begin{equation}
    \mathcal{S} = sgn \left(\sum^{N}_{i=1} \mathcal{F}_i \circ \mathcal{V}_{f_i} \right),
\label{eq:encoding}
\end{equation}
where $f_i$ is the $i$-th feature value,  $\mathcal{V}_{f_i}$ is the corresponding value hypervector, $\circ$ is the Hadamard product, and $sgn(\cdot)$ is the sign function that binarizes the encoded sample hypervector. As a tiebreaker, we set $sgn(0) = 1$.

{\textbf{Training:}} 
Given $K$ classes, the training process is to obtain $K$ class hypervectors, each for one class. The basic training process is to simply average the sample hypervectors within a class:
\begin{equation}
    \mathcal{C}_k = sgn\left(\sum_{\mathcal{S}\in \Omega_k} \mathcal{S}\right)
\label{eq:naive_training}
\end{equation}
where $\Omega_k$ is the set of sample hypervectors in $k$-th class. 
All the class hypervectors are stored in the associative memory (AM). 

More recently, to improve the accuracy, SOTA HDC models have also added re-training as part of the training process \cite{imani2019quanthd,HDC_Survey_Review_IEEE_Circuit_Magazine_2020_9107175}. Concretely, re-training fine tunes the class hypervectors $\mathcal{C}$ derived in Equation~\eqref{eq:naive_training}: if a training sample is mis-classified, it will be given more weights in correct class hypervector and subtracted from the wrong class hypervector. Essentially, re-training will lead to an adjusted centroid for each class.

{\textbf{Inference:}} 
The testing input sample is first encoded in the same way as encoding a training sample. To be distinguished from the training sample hypervector, the testing sample hypervector is  also referred to as a query hypervector. For inference, the query hypervector $\mathcal{S}_q$ is compared with all the class hypervectors fetched from the associative memory. The most similar one  with the lowest Hamming distance indicates the classification result:
\begin{equation}
     {\arg\min_k}\ Hamm(\mathcal{S}_q, \mathcal{C}_k).
    \label{eq:inference_hamm}
\end{equation}
Due to the equivalence of Hamming distance and cosine similarity (appendix), the classification rule in Equation~\eqref{eq:inference_hamm} is equivalent to
\begin{equation}
   {\arg\max_k}\ \mathcal{S}_q^T \mathcal{C}_k,
    \label{eq:inference_cos}
\end{equation}
which essentially transforms the bit-wise comparison to vector multiplication and is instrumental for establishing the equivalence between an HDC model and a neural network.

\section{The Design of \ouralg Classifiers}

In this section, we describe the \ouralg classifier design which exploits hardware-friendly association of low-dimensional vectors for efficient inference. Specifically, like in its HDC counterpart, our \ouralg classifier mimics  brain cognition for hardware efficiency by representing features using vectors and performing inference via vector association \cite{kanerva2009hyperdimensional,HDC_Survey_Review_IEEE_Circuit_Magazine_2020_9107175}. Nonetheless, the existing HDC models rely on randomly-generated hypervectors, which not only limits the accuracy but also results in a large inference latency and resource consumption beyond the capability of tiny devices. Our \ouralg classifier fundamentally differs from them --- it uses vectors with orders-of-magnitude smaller dimensions and optimizes the vectors using a principled approach.

With a slight abuse of notations, we keep using $\mathcal{F}$ and $\mathcal{V}$ to denote  feature and value vectors, respectively, in our \ouralg classifier wherever applicable. But, unlike in an HDC model that has the same hyperdimension of $D$ for all hypervectors, $\mathcal{F}$ and $\mathcal{V}$  can have different and much lower dimensions of $D_{\mathcal{F}}$ and $D_{\mathcal{V}}$, respectively. 

\subsection{Mapping to a Neural Network}

We now divide the \ouralg classification process into two parts --- encoding and similarity checking --- and map them to equivalent neural network operation.

\subsubsection{Encoding}\label{sec:encoding}
The encoding process shown in Equation~\eqref{eq:encoding} binarizes the summed bindings of value vectors and feature vectors. Instead of using random vectors as in the existing HDC models, we explicitly optimize the value and feature vectors by representing the encoder as a neural network and then using a principled training process. Next, we describe the equivalence between the encoder and its neural network counterpart.

\begin{figure}[!t]
  \centering
  \includegraphics[width=\linewidth]{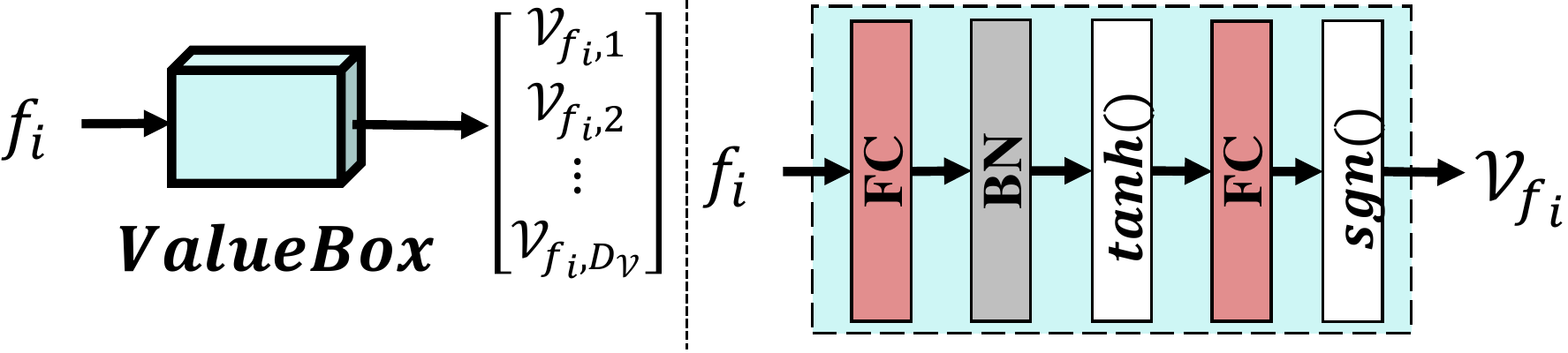}
  \caption{Value mapping. As an example, we use a 2-layer fully connected network as the \textit{ValueBox}. FC: fully-connected, BN: batch normalization.}
  \label{fig:value_mapping}
\end{figure}

{\textbf{Value mapping:}} As shown in Figure~\ref{fig:value_mapping}, $\mathcal{V}_{f_i}$ represents a discretized feature value $f_i\in \{1,\cdots, M\}$ with a certain (bipolar) vector. For the ease of presentation, we refer to this mapping functionality as \textit{ValueBox}. In an HDC model, \textit{ValueBox} essentially assigns a random hypervector to a feature value. Alternatively, one may manually design a \textit{ValueBox}, e.g., representing a value directly using its binary code (say, $9\to \texttt{1001}$) \cite{zhang2021fracbnn}. In our \ouralg design, however, we exploit the strong representation power of neural networks and use a trainable neural network for the \textit{ValueBox}. For example, Figure~\ref{fig:value_mapping} illustrates a simple fully-connected neural network with $tanh$ activation to map a value $f_i$ to a binary value vector $\mathcal{V}_{f_i}$. We will jointly train the \textit{ValueBox} network together with subsequent operators to optimize the inference accuracy.

{\textbf{Element-wise binding:}} 
As shown in Equation~\eqref{eq:encoding}, the encoder binds the feature and value vectors using a Hadamard product. Nonetheless, in our \ouralg classifier, we do not require $D_\mathcal{F} = D_\mathcal{V}$, which makes the Hadamard product inapplicable. Here, we set $D_\mathcal{F}$ as an integer multiple of $D_\mathcal{V}$, i.e., $D_\mathcal{F}/D_\mathcal{V} = n$, for $n\in \mathbb{N}^+$. As a result, a value vector  $\mathcal{V}_{f_i}$ can be stacked for $n$ times in order to have the same dimension as its corresponding feature vector $\mathcal{F}_i$ for Hadamard product. Equivalently, a feature vector can be evenly divided into $n$ parts or sub-vectors, each aligned with the value vector for Hadamard product. Thus, the binding for the $i$-th feature vector and its corresponding value vector can be represented as
\begin{equation}
\begin{bmatrix}
\mathcal{F}_i^1\circ \mathcal{V}_{f_i}\\ 
...\\ 
\mathcal{F}_i^n\circ \mathcal{V}_{f_i}
\end{bmatrix}
= 
\begin{bmatrix}
diag(\mathcal{F}_i^1)\\ 
...\\ 
diag(\mathcal{F}_i^n)
\end{bmatrix} \mathcal{V}_{f_i}.
\end{equation}
Through element-wise binding, the encoder
outputs a sample vector given by
\begin{equation}\label{eqn:encoding_new}
\mathcal{S}=
sgn \left(\sum_{i=1}^N
\begin{bmatrix}
diag(\mathcal{F}_i^1)\\ 
...\\ 
diag(\mathcal{F}_i^n)
\end{bmatrix} \mathcal{V}_{f_i}
\right).
\end{equation}

\begin{figure}[!t]
  \centering
  \includegraphics[width=\linewidth]{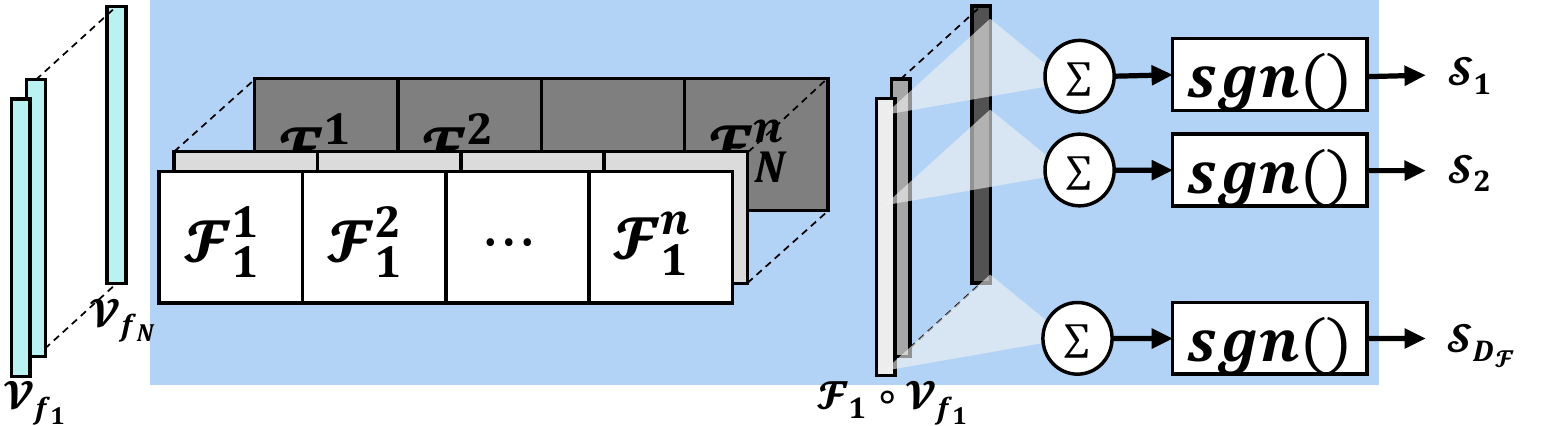}
  \caption{Illustration of a \textit{feature layer} that binds value and feature vectors.}
  \label{fig:bitwise_computing}
\end{figure}

Crucially, the element-wise binding used by the encoder  is equivalently mapped to matrix multiplication in  Equation~\eqref{eqn:encoding_new}. Therefore, it can be represented as a simple BNN whose architecture is  shown in Figure~\ref{fig:bitwise_computing} and referred to as a \emph{feature layer}. Specifically, by stacking $N$ value vectors $\mathcal{V}_{f_i}$ for $i=1,\cdots,N$, the input to the feature layer is a vector with  $N D_{\mathcal{V}}$ elements. Also, the structurally sparse weight matrix $\mathbf{\Theta}$ in the feature layer is the collection of feature vectors $\mathcal{F}_i$ for $i=1,\cdots,N$, with only $ND_{\mathcal{F}}$ bipolar elements in total. By transforming the encoder into an equivalent BNN, we can leverage a principled training process (described in Section~\ref{sec:training}) to optimize the weight matrix $\mathbf{\Theta}$, which in turn leads to optimized feature vectors $\mathcal{F}_i$ with a small dimension instead of random hypervectors used by the existing HDC models.

\begin{figure}[!t]
  \centering
  \includegraphics[width=0.8\linewidth]{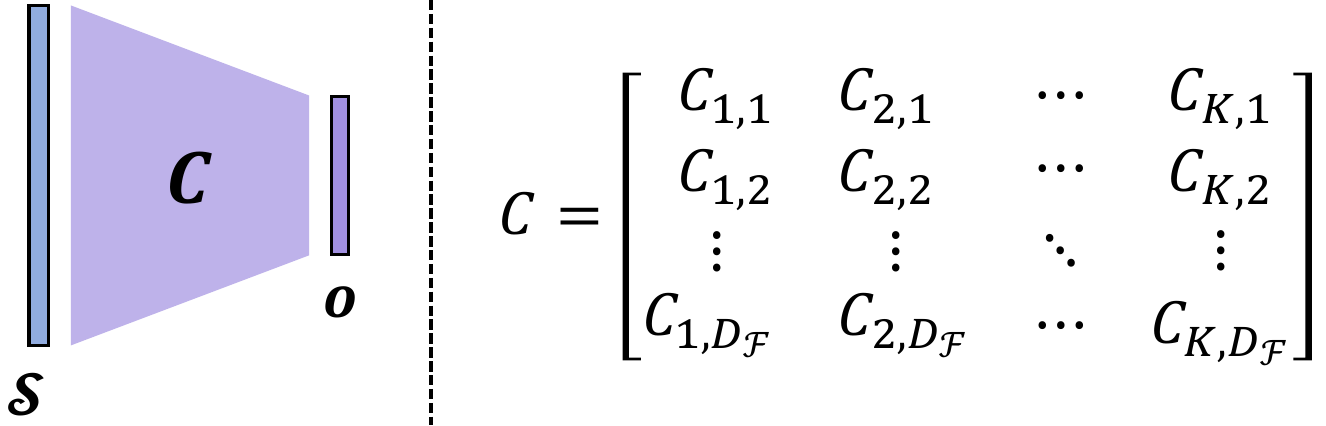}
  \caption{Illustration of a \textit{class layer}. }
  \label{fig:class_layer}
\end{figure}

\subsubsection{Similarity Checking}
In our \ouralg classifier, the classification process involves checking the similarities between a query hypervector and all class hypervectors. Specifically, as shown in Equations~\eqref{eq:inference_hamm} and \eqref{eq:inference_cos}, the similarity check is equivalent to matrix calculation, which can also be mapped to the operation in a BNN. Hence, we transform Equation~\eqref{eq:inference_cos} into a \textit{class layer} as shown in Figure \ref{fig:class_layer}. The input to the class layer is the sample vector, which is also the output of the preceding feature layer. The weight of the class layer is a $D_{\mathcal{F}}\times K$ matrix that represents the collection of all the class vectors $\mathcal{C}_k$ for $k=1,\cdots,K$. The output of the class layer includes $K$ products, and class $k$ with the maximum product  $\mathcal{S}^T \mathcal{C}_k$ is chosen as the classification result.

\subsubsection{End-to-end mapping.}
\begin{figure}[!t]
  \centering
  \includegraphics[width=\linewidth]{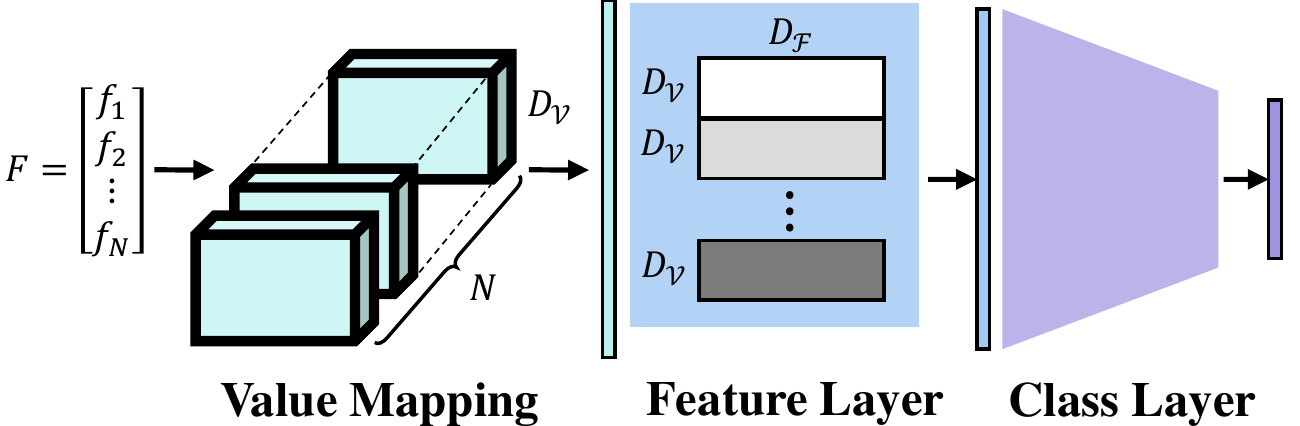}
  \caption{Mapping an \ouralg classifier to a neural network.}
  \label{fig:integration}
\end{figure}

By integrating different stages of the  \ouralg classification pipeline, we now construct an end-to-end neural network shown in Figure~\ref{fig:integration}. With both non-binary and binary weights, the neural network achieves the same function as our \ouralg classifier, and this equivalence allows a principled training approach to optimize the weights.

Specifically, in the equivalent neural network, each of the $N$ feature values of an input sample is first fed to a \textit{ValueBox}, which is a non-binary neural network and outputs bipolar value vectors. Note that a single \textit{ValueBox} is shared by all features to keep the model size small, while our design can easily generalize to different \textit{ValueBoxes} for different features at the expense of increasing the model size (in particular, the size of item memory). Then, the $N$ value vectors enter a feature layer, which is a structurally sparse BNN as described in Section~\ref{sec:encoding} and outputs a sample vector. Finally, the sample vector goes through a class layer, based on which we can decide the classification result.

\textbf{SOTA HDC models:} 
Besides the obvious drawback of ultra-high dimension, we can clearly see another major drawback by casting an existing HDC model into our equivalent neural network. Concretely, the \textit{ValueBox} outputs and the feature layer weights corresponding to an HDC model are essentially \emph{randomly} generated, and even the weights in the class layer (i.e., $K$ class hypervectors in an HDC model) are obtained by using simple averaging methods in conjunction with heuristic re-training \cite{HDC_Survey_Review_IEEE_Circuit_Magazine_2020_9107175,kanerva2009hyperdimensional,HDC_BinaryLearning_UCSD_DATE_2019_8714821}. Thus, the inference accuracy in the existing
HDC models are highly sub-optimal. 

\subsection{Training \ouralg Classifiers}
\label{sec:training}
To address the fundamental drawbacks of existing HDC models and maximize the accuracy with a much smaller model size, we can optimize the weights in its equivalent neural network that has both non-binary weights (in the \textit{ValueBox}) and binary weights (in the feature layer and class  layer). Specifically, following SOTA training methods for BNNs \cite{liu2021adam}, we use \textit{Adam} with weight decay  as the optimization method and consider softmax activation with \textit{CrossEntropy} as the loss function. Due to the equivalence of Hamming distance and cosine similarity metrics for binary vectors, classification based on the largest softmax probability (or equivalently,  ${\arg\max_k} \mathcal{S}_q^T \mathcal{C}_k$) is equivalent to classification based on the minimum normalized Hamming distance. Additionally, while CrossEntropy is commonly used for classification tasks, we can also use other loss functions such as hinge loss \cite{DNN_Book_Goodfellow-et-al-2016}. For training, we set the learning rate by starting with a large value (such as 0.001) with decaying linearly.

After training, the vectors used in our \ouralg classifier can be extracted as follows for testing.

\textbf{\circled{1}} The value vectors $\mathcal{V}$ can be extracted from the \textit{ValueBox} by recording the output corresponding to each possible input. For example, the value vector ($\mathcal{V}_f$) for a certain feature value $f$ is $ValueBox(f)$. Due to $D_\mathcal{F}/D_{\mathcal{V}} = n$ for Hadamard product, the value vector is stacked $n$ times in the encoder to align with the dimension of feature vectors. Thus, the non-binary weights in the \textit{ValueBox}   are not utilized for inference and can be discarded once the value vectors are extracted.

\textbf{\circled{2}} The feature vectors $\mathcal{F}$ are extracted from the weight matrix in the {feature layer}. For the $i$-th feature, the feature vector $\mathcal{F}_i$ can be extracted from the corresponding values in the $i$-th weight matrix block:
\begin{equation*}
\begin{bmatrix}
 \mathcal{F}_{i,1}^1 & 0 & ... & 0 & \cdots  & \mathcal{F}_{i,1}^n & 0 & ... & 0\\
 0 & \mathcal{F}_{i,2}^1 & ... & 0 & \cdots  & 0 & \mathcal{F}_{i,2}^n & ... & 0\\
 \vdots  & \vdots  & \ddots  & \vdots  & \cdots  & \vdots  & \vdots  & \ddots  & \vdots \\
 0 & 0 & ... & \mathcal{F}_{i,D_\mathcal{V}}^1 & \cdots  & 0 & 0 & ... &\mathcal{F}_{i,D_\mathcal{V}}^n
\end{bmatrix}
\end{equation*}
such that $\mathcal{F}_i$ is composed of $[\mathcal{F}_{i}^1, \mathcal{F}_{i}^2, \cdots, \mathcal{F}_{i}^n]$ with dimension $D_\mathcal{F}$, where $\mathcal{F}_{i}^j=[\mathcal{F}_{i,1}^j,\cdots,\mathcal{F}_{i,D_\mathcal{V}}^j]$ for $j=1,2,\cdots,n$.

\textbf{\circled{3}} As shown in the right side of Figure~\ref{fig:class_layer}, the class vectors $\mathcal{C}$ can be directly extracted from the weight matrix in \textit{class layer}. 

Like in Figure~\ref{fig:binary_HDC}, the extracted value and feature vectors are stored in the item memory for encoding, and the class vectors
are stored in the associative memory for similarity checking.

\subsection{Inference}
For inference, our \ouralg classifier still follows the encoding and similarity checking process as described in Section~\ref{sec:background}. Specifically, each feature value of an input sample is first mapped to a value vector, which is then combined together with the corresponding feature vector to form  a query sample vector. The query vector is compared with the class vectors for similarity checking and yielding the classification results. In most BNNs, the fully-connected layers still use non-binary weights, which can slow down the inference process on tiny devices \cite{zhang2021fracbnn}. By contrast, although the offline training process involves non-binary weights in the \textit{ValueBox} neural network, the inference process of our \ouralg classifier is fully binary by utilizing bit-wise operation and association for classification. 

\section{Performance Evaluation}
In this section, we evaluate the performance of our \ouralg classifier  on different datasets and highlight that it offers an overwhelming advantage over the SOTA HDC models: better accuracy and orders-of-magnitude smaller dimension.

\subsection{Experimental Setup}
Like in the existing HDC research \cite{HDC_BinaryLearning_UCSD_DATE_2019_8714821,HDC_Survey_Review_IEEE_Circuit_Magazine_2020_9107175}, we select four application scenarios for inference on tiny devices: computer vision (MNIST \cite{726791} and Fashion-MNSIT \cite{xiao2017/online}), human activity (UCIHAR \cite{anguita2013public}), voice recognition (ISOLET \cite{isolet_54}), and cardiotocography (CTG \cite{cardiotocography_193}). Each feature value is normalized to $[0,255]$ and quantized to an 8-bit integer. The configurations for each dataset is shown in Table~\ref{tab:configuration}. We compare \ouralg\ with the SOTA HDC model with re-training \cite{imani2019quanthd}, and the basic HDC model without re-training \cite{HDC_Survey_Review_IEEE_Circuit_Magazine_2020_9107175}. As reported in \cite{imani2019quanthd}, the HDC accuracy has significant reduction when the hypervector dimension is lower than $8,000$. Thus, we use $D=8,000$ for both SOTA and basic HDC. For training, all the experiments are executed in Python with Tesla V100 GPU. For the inference, we also build a hardware acceleration platform on a Zynq UltraScale+ ZU3EG FPGA embedded on the Ultra96 evaluation board.

\subsection{Hyperparameter Selection}

\subsubsection{Value mapping} 
\begin{figure}[!t]
\centering
\subfigure{
	    \centering
		\includegraphics[width=0.55\linewidth]{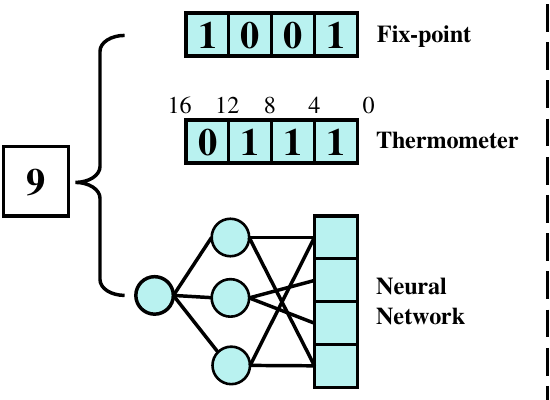}}
\subfigure{
	    \centering
		\includegraphics[width=0.35\linewidth]{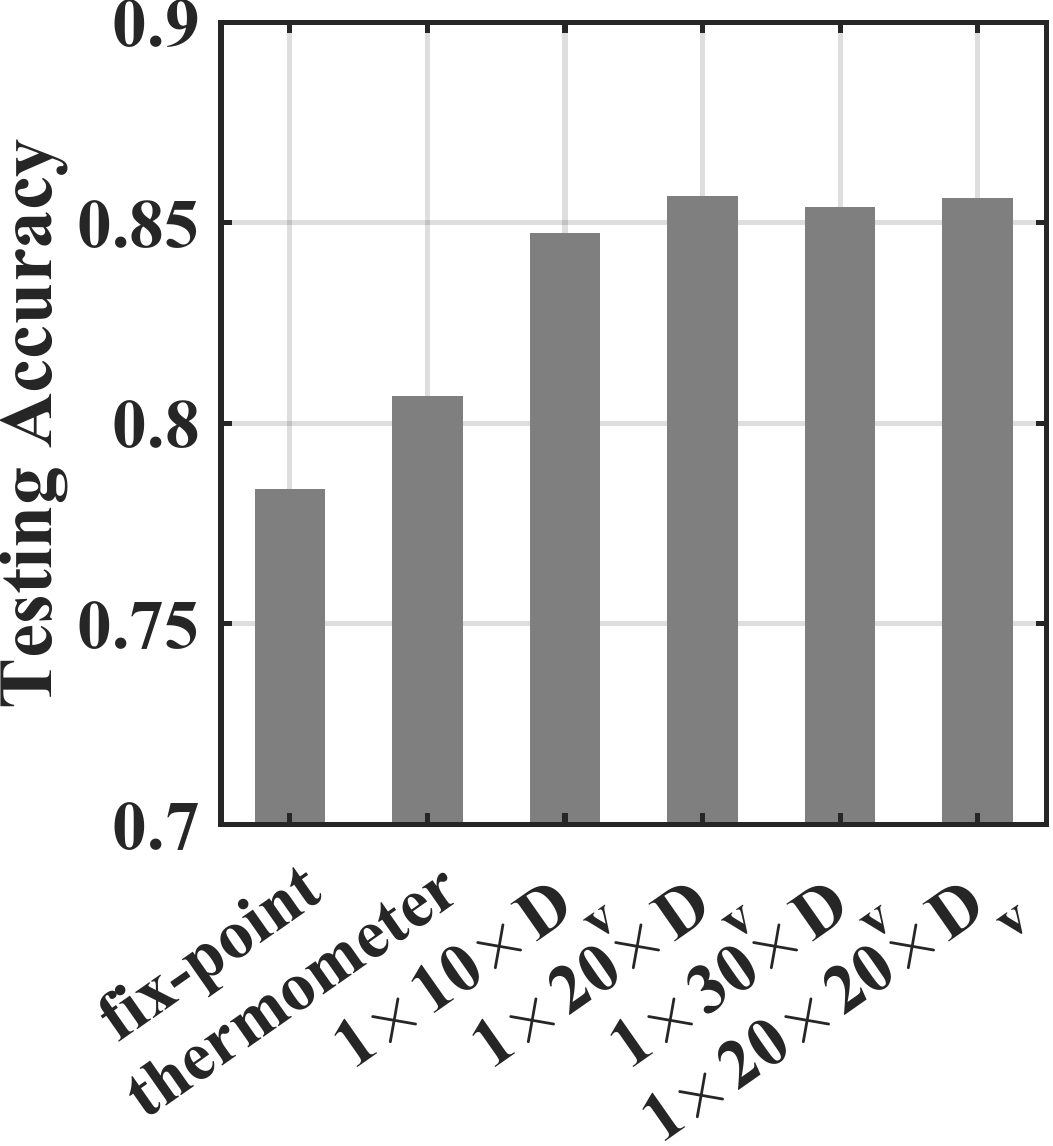}}
\caption{Different \textit{ValueBoxes}. The testing accuracies  are presented on the right side with Fashion-MNIST  \cite{xiao2017/online}.}
\label{fig:training}
\end{figure}

We test different neural networks for the \textit{ValueBox} and also compare them against two manual designs (i.e.,  fix-point encoding and thermometer encoding \cite{zhang2021fracbnn}, as illustrated in Figure~\ref{fig:training}). We set $D_\mathcal{V}=8$ to make fix-point encoding and thermometer encoding capable of representing 256 values. Considering the Fasion-MNIST dataset, we see that our neural networks can discover better \textit{ValueBoxes} than manual designs to achieve higher accuracy. On the other hand, it has no significant differences by varying the network size, such as $1\times10\times D_\mathcal{V}$ vs. $1\times20\times D_\mathcal{V}$. In our experiments, we will use $1\times 20\times D_{\mathcal{V}}$ for the neural network in the \textit{ValueBox}.

\subsubsection{Dimensions} 
\begin{figure}[!t]
  \centering
  \includegraphics[width=0.8\linewidth]{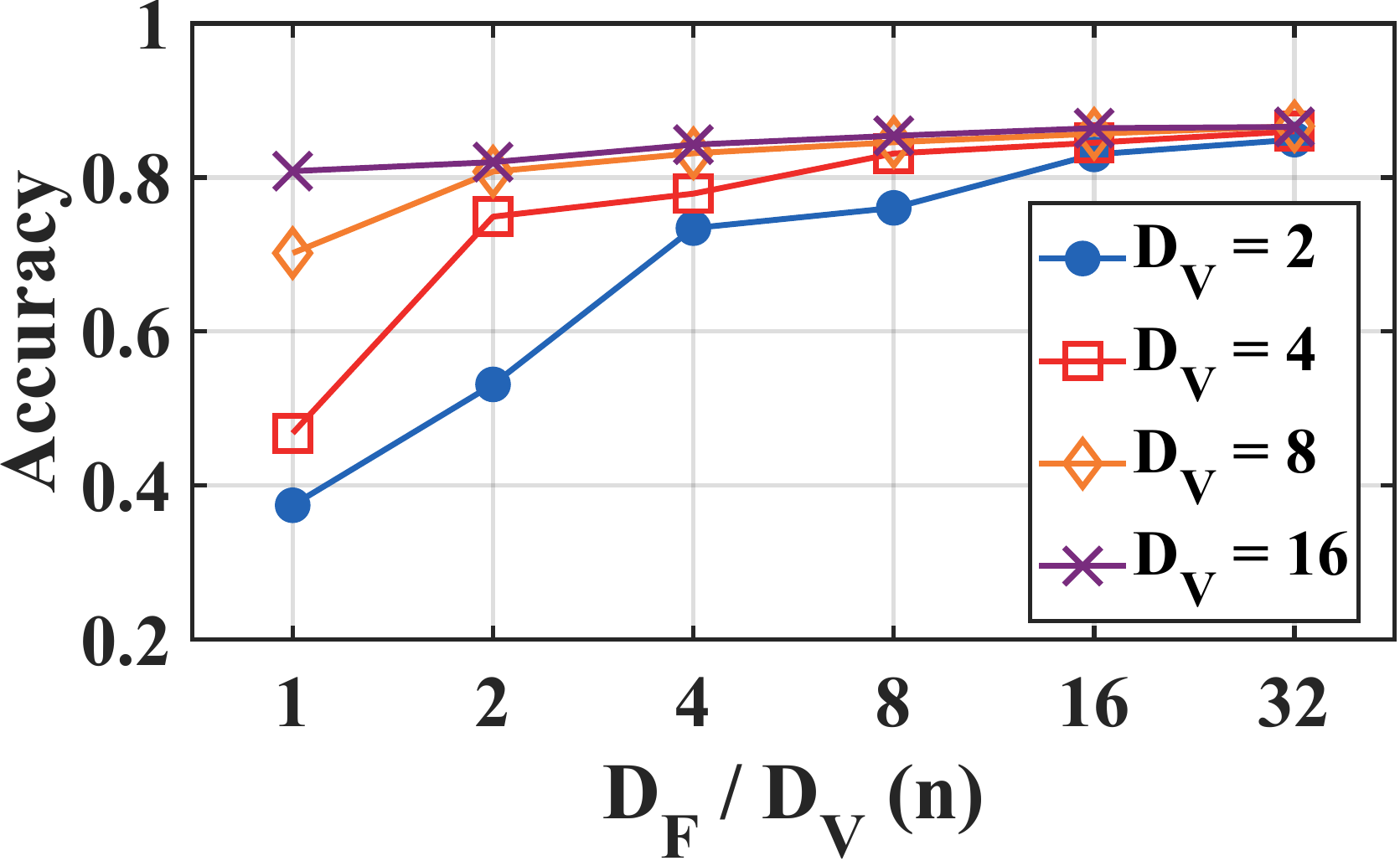}
  \caption{The inference accuracy with different $D_\mathcal{V}$ and $D_\mathcal{F}$ on Fashion-MNIST. Initial learning rate is $1\times 10^{-3}$, weight decay is $1\times 10^{-4}$, and learning rate will decay by half every 5 iterations.}
  \label{fig:dimension_selection}
\end{figure}

The dimensions
$D_\mathcal{V}$ and $D_\mathcal{F}$ used in our \ouralg classifier are important hyperparameters. As a case study, we use Fashion-MNIST to evaluate the impact of $D_\mathcal{V}$ and $D_\mathcal{F}$  on the  accuracy. By setting different $n=D_\mathcal{F}/D_\mathcal{V}$, we also vary $D_\mathcal{V} = 2, 4, 8, 16$, as presented in Figure~\ref{fig:dimension_selection}. In general, a higher $D_\mathcal{V}$ retains richer information about the input, thus  yielding a higher accuracy. Nonetheless, even with $D_\mathcal{V} = 2$ or $D_\mathcal{V} = 4$, we can still achieve a reasonably good accuracy by increasing $D_\mathcal{F}$. In all cases, our dimensions are orders-of-magnitude smaller than the dimensions of 8,000 or higher in existing HDC models. 

\begin{table}[!t]
\caption{Configuration of our \ouralg classifier 
for each dataset. The batch size is 64.}
\resizebox{\linewidth}{!}{
\begin{tabular}{l|ccccc}
\toprule
\multicolumn{1}{c|}{\textbf{Dataset}} & \multicolumn{1}{c}{\textbf{N}} & \multicolumn{1}{c}{\textbf{\# of (train, test, class)}} & \multicolumn{1}{c}{\textbf{$D_\mathcal{V}$, $D_\mathcal{F}$}} & \multicolumn{1}{c}{\textbf{LR}$^{\mathrm{1}}$} & \multicolumn{1}{c}{\textbf{WD}$^{\mathrm{2}}$} \\ \hline
MNIST & 784 & (60000, 10000, 10) & 4, 64 & 0.0001 & 0 \\
Fashion-MNIST & 784 & (60000, 10000, 10) & 4, 64 & 0.0002 & 0.00001 \\
UCIHAR & 561 & (7352, 2947, 6) & 4, 128 & 0.001 & 0.0001 \\
ISOLET & 617 & (6328, 1559, 26) & 4, 128 & 0.005 & 0.0001 \\
CTG & 21 & (1701, 425, 3) & 4, 64 & 0.008 & 0.0001\\
\bottomrule
\multicolumn{6}{l}{$^{\mathrm{1}}$ LR = Learning Rate \space $^{\mathrm{2}}$ WD = Weight Decay}
\end{tabular}}
\label{tab:configuration}
\end{table}

\subsection{Inference Accuracy}
In Table \ref{tab:acc_compare}, we evaluate the inference accuracy that \ouralg\ can achieve compared to both basic and SOTA HDC models. The result shows that \ouralg\ can outperform existing HDC models that use random value/feature vectors and heuristically generate class vectors. While retraining improves the inference accuracy against the basic HDC, the hypervector dimension must be as large as 8,000 to prevent accuracy degradation. By contrast, our \ouralg classifier reduces the dimension significantly, without introducing any extra cost during the inference process.

\begin{table}[!t]
\caption{Inference accuracy (\%, in $mean^{\pm std}$).}
\resizebox  {\linewidth}{!}{
\begin{tabular}{l|ccccc}
\toprule
\multicolumn{1}{c|}{\textbf{Classifier}} & \multicolumn{1}{c}{\textbf{MNIST}} & \multicolumn{1}{c}{\textbf{Fashion-MNIST}} & \multicolumn{1}{c}{\textbf{UCIHAR}} & \multicolumn{1}{c}{\textbf{ISOLET}} & \multicolumn{1}{c}{\textbf{CTG}} \\ \hline
LDC & $\textbf{92.72}^{\pm 0.18}$ & $\textbf{85.47}^{\pm 0.29}$ & $\textbf{92.56}^{\pm 0.41}$ & $\textbf{91.33}^{\pm 0.50}$ & $\textbf{90.50}^{\pm 0.46}$ \\
Basic HDC & $79.35^{\pm 0.03}$ & $69.11^{\pm 0.03}$ & $89.17^{\pm 0.16}$  & $85.90^{\pm 0.25}$ & $71.35^{\pm 0.76}$ \\
SOTA HDC \cite{imani2019quanthd} & $87.38^{\pm 0.21}$ & $79.24^{\pm 0.52}$ & $90.31^{\pm 0.06}$ & $90.66^{\pm 0.31}$ & $89.51^{\pm 0.43}$\\
\bottomrule
\end{tabular}}
\label{tab:acc_compare}
\end{table}

\begin{table*}[!t]
\caption{Hardware acceleration on inference with two selected datasets. The system frequency is 200MHz.}
\resizebox  {\linewidth}{!}{
\begin{tabular}{cl|cccccc}
\toprule
\multicolumn{1}{c}{\textbf{Dataset}} & \multicolumn{1}{c|}{\textbf{Name}} &
\multicolumn{1}{c}{\textbf{Accuracy (\%)}} &
\multicolumn{1}{c}{\textbf{Platform}} & \multicolumn{1}{c}{\textbf{Model Size (KB)}} & \multicolumn{1}{c}{\textbf{Latency ($\mu$s)}} & \multicolumn{1}{c}{\textbf{(LUT, BRAM, DSP)}} & \multicolumn{1}{c}{\textbf{Energy (nJ)}} \\ \hline
\multicolumn{1}{c}{\multirow{3}{*}{\begin{tabular}[c]{@{}l@{}} \textbf{MNIST} \end{tabular}}} &
LDC & 92.72 & Zynq UltraScale+ & \textbf{6.48} & \textbf{3.99} & \textbf{(745, 5, 1)} & \textbf{64} \\
 & SOTA HDC  & 87.38 & Zynq UltraScale+ & 1050 & 499 & (768, 178, 1) & 36926 \\
 & FINN \cite{umuroglu2017finn} & 95.83 & Zynq-7000 & 600 & 240 & (5155, 16, -) & 96000\\\hline
\multicolumn{1}{c}{\multirow{3}{*}{\begin{tabular}[c]{@{}l@{}} \textbf{CTG} \end{tabular}}} &
LDC & 90.50 & Zynq UltraScale+ & \textbf{0.32} & \textbf{0.14} & \textbf{(345, 3, 1)} & \textbf{0.945} \\
& SOTA HDC & 89.51 & Zynq UltraScale+ & 280 & 16.88 & (362, 9, 1) & 169 \\
& Compressed HDC \cite{basaklar2021hypervector} & $\sim$82 & Odroid XU3 & 45.1 & 90 & NA & 6300 \\
\bottomrule
\end{tabular}}
\label{tab:hardware}
\end{table*}

\subsection{Hardware Acceleration}
\subsubsection{Hardware design}
For hardware implementation, bipolar values $\{1,-1\}$ are represented as binary values $\{0,1\}$, respectively. The multiplication and accumulation on bipolar values are realized using \textit{XOR} and \textit{popcount} which is constructed using tree adders, on the binary representation \cite{liang2018fp}. 

The existing HDC models typically exploit full parallelism for acceleration. For example, given $N$ features,  the FPGA prepares $N$ identical hypervector multiplication blocks to encode all the features simultaneously, incurring a high resource expenditure (e.g., over $10^5$ LUTs, 100 BRAMs, and 800 DSPs in SOTA FPGA acceleration \cite{imani2021revisiting}). Nonetheless, this design does not fit into tiny devices, for which we must limit the resource utilization.

\begin{figure}[!t]
  \centering
  \includegraphics[width=0.85\linewidth]{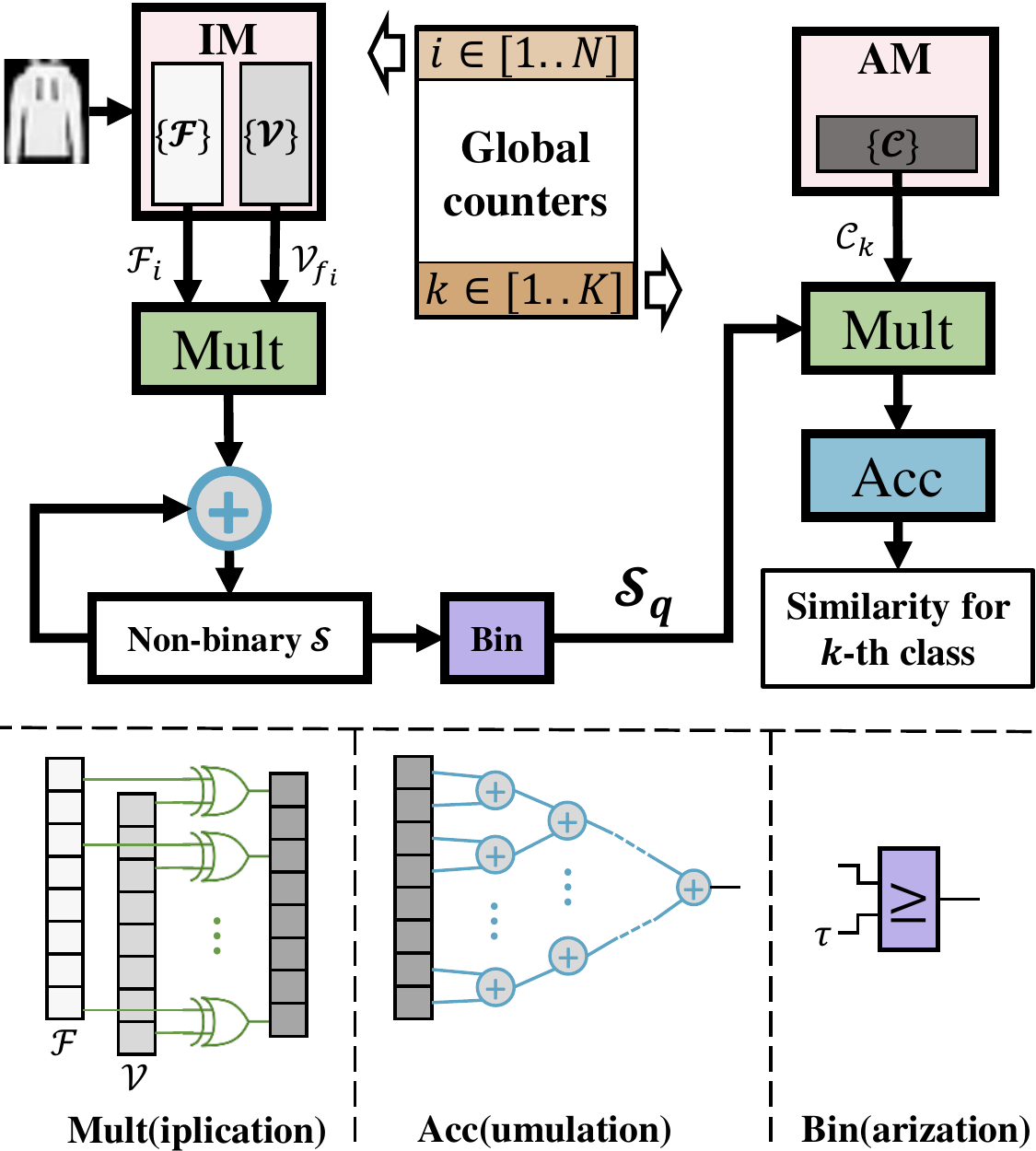}
  \caption{The hardware acceleration for \ouralg\ on FPGA. IM: Item Memory. AM: Associative Memory.}
  \label{fig:acceleration}
\end{figure}

We design a pipeline structure for  feature encoding, rather than in parallel, to fit into  tiny devices. The  structure is demonstrated in Figure~\ref{fig:acceleration}. Only one vector multiplication block and several BRAMs are required. Although the encoding time will increase to $N+1$ cycles, the latency is still on a microsecond scale. Subsequently, one adder is utilized to accumulate the multiplied vectors, followed by a threshold ($\tau = N/2$) comparator to binarize the encoding output. For checking similarity, we also use a pipeline structure for comparison on all class vectors. Then, a tree adder along the vector dimension is used for Hamming distance calculation. Finally, we transfer the Hamming distances back to the CPU to execute the $\arg\min()$ function for classification.\footnote{As an alternative, the $argmin()$ function can also be integrated on FPGA with a small overhead.}

\subsubsection{Results}
In Table~\ref{tab:hardware}, we show the efficiency results. As we focus on tiny devices, we limit the resource utilization, e.g., $<1$MB model size and $<10$k LUT utilization. To evaluate the existing HDC models, for fair comparison, we also implement the SOTA HDC classifier \cite{imani2019quanthd} with $D=8,000$ using our acceleration designs. Moreover, we choose two other lightweight models for comparison: a compressed HDC model \cite{basaklar2021hypervector} that uses a small vector but has non-binary weights and per-feature \textit{ValueBoxes} founded using evolutionary search; and SFC-fix with FINN \cite{umuroglu2017finn} that applies a 3-layer binary MLP on FPGA. For these models, we only report their results available for our considered datasets. Like in the literature \cite{imani2021revisiting,imani2019quanthd}, the results are measured for model inference only, excluding data transmission between the FPGA and CPU.

For the MNIST dataset, by reducing the dimension of SOTA HDC models by $125$ times, the model size of \ouralg\ is only 6.48 KB. Further, the low dimension can benefit the resource utilization and execution time. For the SOTA HDC model with $D=8,000$, the number of BRAMs increases greatly to store all hypervectors, but other resources such as LUT and DSP do not increase dramatically due to  sequential execution to account for tiny devices; consequently, the latency increases by $125$ times. For  energy consumption, the result shows that our \ouralg classifier has the lowest, because of the very low resource utilization and short latency. For the CTG dataset, the results are even more impressive as shown in Table~\ref{tab:hardware}.

\subsection{Robustness Against Random Bit Errors}
\begin{figure}[!t]
  \centering
  \includegraphics[width=0.8\linewidth]{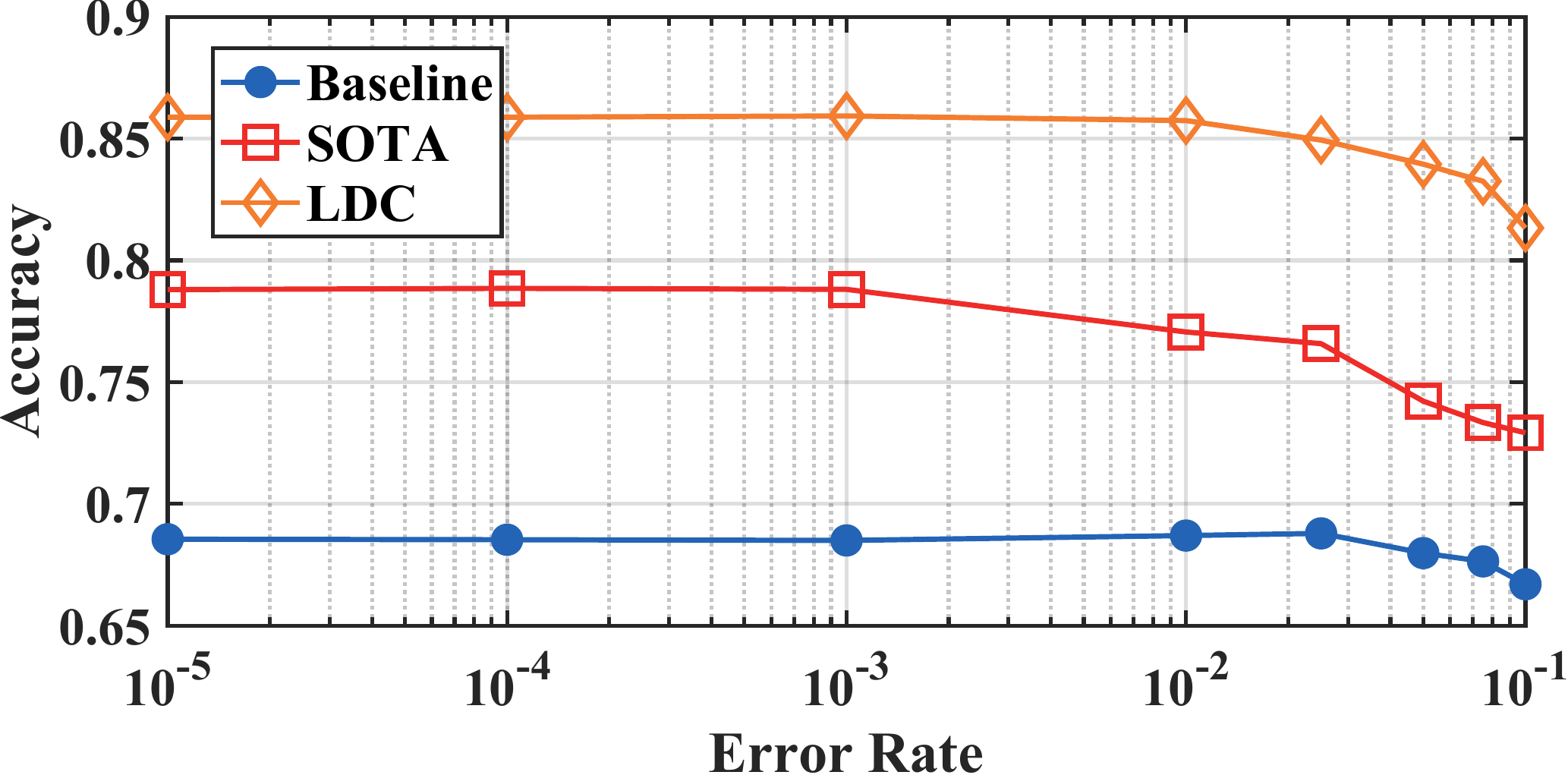}
  \caption{Bit error robustness of different models for Fashion-MNIST dataset. The accuracy is averaged over five different runs for each case.}
  \label{fig:robustness}
\end{figure}
While inefficient, 
a natural byproduct of the hyperdimensionality of HDC classifiers 
is the robustness against random hardware bit errors \cite{HDC_BitFlipping_UCI_ICCAD_2021,HDC_FuzzTesting_DAC_2021_9586169}.
By contrast, our \ouralg classifier reduces the dimension by orders of magnitude
and hence might become less robust against random bit errors.
Here, we show in Figure~\ref{fig:robustness}
the robustness analysis for both \ouralg and HDC classifiers. 
Specifically, we inject bit errors in the associative memory under various bit error rates, and assess the inference accuracy caused by the injected bit errors. We
see that our \ouralg classifier has
the highest accuracy and also achieves comparable robustness on a par with the HDC counterparts.
The counter-intuitive results can be explained by the fact that, although
our \ouralg classifier
significantly reduces  dimensions, the information is still spread uniformly within each compact 
vector (i.e., different bits are equally important), thus exhibiting robustness against random bit errors.

Note finally that
we can also run multiple \ouralg classifiers and apply the majority rule for better robustness.
 Due to the orders-of-magnitude dimension reduction,  the total size of multiple
\ouralg classifiers is still much less than that of a single HDC classifier.

\section{Related Work}

The set of applications with HDC classifiers have been proliferating,  include language classification \cite{HDC_LowPower_LanguageRecognition_UCSD_DesignTest_2017_8012434}, image classification \cite{HDC_Survey_Review_IEEE_Circuit_Magazine_2020_9107175,HDC_TaskProjected_MultiTask_Learning_2020_10.1007/978-3-030-49161-1_21}, emotion recognition based on physiological signals \cite{HDC_EmotionRecognition_AICAS_2019_8771622}, distributed fault isolation in power plants \cite{HDC_IndustrialSystemsPowerPlant_IEEE_Access_2018_kleyko2018hyperdimensional},  gesture recognition for wearable devices \cite{HDC_LowPower_EMG_Gesture_Classification_Trans_2019_8704957}, seizure onset detection  \cite{HDC_SeizureOnsetDetection_Trans_2019_8723166}, and robot navigation \cite{HDC_Robot_Learning_UMD_2019_Mitrokhineaaw6736}.  

Nonetheless, the training process for existing HDC classifiers is
mostly heuristics-driven and often relies on random hypervectors for encoding, resulting in low inference accuracy \cite{HDC_Survey_Review_IEEE_Circuit_Magazine_2020_9107175}.
In many cases, even a well-defined loss function is lacking 
for training HDC classifiers.
More recently,
\cite{HDC_LearningHolographicReducedRepresentations_ganesan2021learning}
learns  a neural network model
based on non-binary hypervector representations, and
\cite{HDC_GeneralizedLVQ_RNN_HDC_diao2021generalized} uses generalized
learning vector quantization (GLVQ) to optimize
the class vectors.
But, they both consider
hyperdimensional representations based on random hypervectors, thus resulting in an overly large model size.

In addition, the ultra-high dimension is another fundamental drawback of HDC models. Simply reducing the dimension without fundamentally changing the HDC design
can dramatically decrease the inference accuracy 
A recent study \cite{basaklar2021hypervector} reduces the dimension but considers non-binary HDC that is unfriendly to hardware acceleration. Moreover, it uses different sets of value vectors for different features, and hence results in a large model size (Table~\ref{tab:hardware}). Last but not least, its training process is still heuristic-driven as in the existing HDC models. By sharp contrast, our \ouralg classifier is optimized based on a principled training approach guided by a loss function, and offers an overwhelming advantage over the existing HDC models, in terms of accuracy, model size, inference latency, and energy consumption.
 
Our \ouralg classifier is relevant to but also differs significantly from BNNs which use binary weights to speed up inference  \cite{zhang2021fracbnn,BNN_XNOR_Net_ECCV_2016_10.1007/978-3-319-46493-0_32,
BNN_Survey_arXiv_2021_DBLP:journals/corr/abs-2110-06804}. To avoid information loss, non-binary weights are still utilized in the early stage of typical BNNs \cite{BNN_XNOR_Net_ECCV_2016_10.1007/978-3-319-46493-0_32, BNN_Survey_arXiv_2021_DBLP:journals/corr/abs-2110-06804} which may not be supported by tiny devices, whereas the inference of our \ouralg classifier is fully binary and follows a brain-like cognition process. More recently, \cite{zhang2021fracbnn} manually designs the value mapping from raw features to binary features. In our design, we use a neural network to automatically learn the mapping which, as shown in Figure~\ref{fig:training},  outperforms manual designs in terms of accuracy.

\section{Conclusion}
In this paper, we address the fundamental drawbacks of existing HDC models and propose an \ouralg alternative. We map our \ouralg classifier into an equivalent neural network, and optimize the model using a principled training approach. We run experiments to evaluate our \ouralg classifier, and also implement different models on an FPGA platform for acceleration. The results highlight that our \ouralg classifier has an overwhelming advantage over the existing HDC models and is particularly suitable for inference on tiny devices.

\appendix
\section*{Appendix}
We show the equivalence between the (normalized) Hamming distance and \textit{cosine} similarity on two bipolar vectors. The normalized Hamming distance and \textit{cosine} similarity are defined as: $Hamm(\mathcal{H}_1, \mathcal{H}_2) = \frac{|\mathcal{H}_1\neq \mathcal{H}_2|}{D}$, and     $cosine(\mathcal{H}_1, \mathcal{H}_2) = \frac{\mathcal{H}_1^T \mathcal{H}_2}{\|\mathcal{H}_1\|\ \|\mathcal{H}_2\|}$, where $|\mathcal{H}_1\neq \mathcal{H}_2|$ denotes the number of different bits in $\mathcal{H}_1$ and $\mathcal{H}_2$,  $\|\mathcal{H}_1\|$ and $\|\mathcal{H}_2\|$ denote the $l_2$ norms of $\mathcal{H}_1$ and $\mathcal{H}_2$, respectively. Since vectors are in bipolar $\{1, -1\}$, we have $\mathcal{H}_1^T \mathcal{H}_2 = (|\mathcal{H}_1 = \mathcal{H}_2| - |\mathcal{H}_1 \neq \mathcal{H}_2|)$. Plus the fact that $\|\mathcal{H}_1\|\ \|\mathcal{H}_2\| = D$ and $(|\mathcal{H}_1\neq \mathcal{H}_2| + |\mathcal{H}_1= \mathcal{H}_2|) = D$, we can conclude $cosine(\mathcal{H}_1, \mathcal{H}_2) = \frac{D-2|\mathcal{H}_1\neq \mathcal{H}_2|}{D}
   =1-2 Hamm(\mathcal{H}_1, \mathcal{H}_2)$.



\end{document}